\documentclass[runningheads]{llncs}
\usepackage{amsmath} 
\usepackage{amsxtra} 
\usepackage{amssymb} 
\usepackage{graphicx}
\usepackage{mathtools}
\usepackage{multirow}
\usepackage[]{algorithm}
\usepackage{algorithmic}
\usepackage{comment}
\DeclareMathOperator*{\argmax}{arg\,max}
\DeclareMathOperator*{\argmin}{arg\,min}

\newlength\myindent
\newcommand\bindent{%
  \begingroup
  \setlength{\itemindent}{\myindent}
  \addtolength{\algorithmicindent}{\myindent}
}
\newcommand\eindent{\endgroup}
\begin{document}
\title{Analysis and Improvement of Adversarial Training in DQN Agents With Adversarially-Guided Exploration}
\titlerunning{Adversarially-Guided Exploration for Robust DQN}
%
\author{Vahid Behzadan\inst{1} \and
William Hsu\inst{1}\\
\authorrunning{V. Behzadan et al.}
%
\institute{Kansas State University}
\email{\{behzadan, bhsu\}@ksu.edu}}
%
\maketitle              
\begin{abstract}
 This paper investigates the effectiveness of adversarial training in enhancing the robustness of Deep Q-Network (DQN) policies to state-space perturbations. We first present a formal analysis of adversarial training in DQN agents and its performance with respect to the proportion of adversarial perturbations to nominal observations used for training. Next, we consider the sample-inefficiency of current adversarial training techniques, and propose a novel Adversarially-Guided Exploration (AGE) mechanism based on a modified hybrid of the $\epsilon$-greedy algorithm and Boltzmann exploration. We verify the feasibility of this exploration mechanism through experimental evaluation of its performance in comparison with the traditional decaying $\epsilon$-greedy and parameter-space noise exploration algorithms.

\end{abstract}
\section{Introduction}
Recent studies have established the brittleness of Deep Reinforcement Learning (DRL) policies to variations in the state space\cite{rajeswaran2017towards}. This can be attributed to failure in the generalization of the policy with respect to input features\cite{zhang2018dissection}. Consequently, many of the proposed techniques for enhancement of such brittleness are based on the idea of regularization. As a recent survey of literature on defensive techniques illustrates \cite{behzadan2018faults}, a major emphasis in such techniques is on adversarial training \cite{pinto2017robust}, which is in effect a regularization technique based on data augmentation. In this paper, we first present an analysis of adversarial training in Deep Q-Network (DQN) agents\cite{mnih2015human}, and its effectiveness with respect to the proportion of adversarial perturbations used for training. Next, we establish the sample-inefficiency of current adversarial training techniques, and develop a novel adversarially-guided exploration mechanism based on a modified hybrid of the $\epsilon$-greedy and Boltzmann exploration techniques \cite{sutton1998reinforcement}, and evaluate its performance in comparison with the traditional decaying $\epsilon$-greedy and parameter-space noise exploration\cite{fortunato2017noisy} algorithms. 

\section{Limits of Adversarial Training}
In this section, we analyze the effectiveness of training a DRL agent with experiences generated through an adversarial interaction. We consider an adversary constrained to a probabilistic budget $P(attack)$, which is the probability of perturbing any state $s'_t\leftarrow s_t+ \delta$ such that the approximated policy at the $i$th iteration of training ($\pi_i$) produces an incorrect action, i.e., $\pi_i(s'_t) \neq \pi_i(s_t)$. We also consider two types of adversarial objectives, one is the \emph{state-neutral} adversary, which imposes the perturbation so that the resulting $s'_t$ induces any action other than $\pi_i(s_t)$. The second type type of adversary we consider is the \emph{targeted} adversary, which crafts $s'_t$ such that the induced action is the worst possible choice, i.e., $\pi_i(s'_t) = \argmin_a Q_i(s, a)$. We assume that the adversary is always successful in crafting the desired perturbations.

We begin the analysis by noting the effect of such perturbations on the composition of the experience replay memory. For any state $s_t$, two types of experiences may be recorded. One represents the nominal (i.e., unperturbed) experiences, denoted by:
\begin{equation}
    \langle s_t, a_t = \pi_i(s_t), s_{t+1}, r(s_t, a_t, s_{t+1})\rangle
\end{equation}
The second type are experiences in which $s_t$ is the result of perturbing another state, i.e., $s_t \leftarrow s'_t + \delta$. Such adversarial experiences are denoted by:
\begin{equation}
    \langle s_t, a_t = \pi_i(s_t), s'_{t+1}, r(s_t, a_t, s'_{t+1})\rangle
\end{equation}
Hence, the expected TD-error of state $s_t$ in each iteration $i+1$ of training is given by:
\begin{equation}
\begin{aligned}
  \mathbb{E}[Err_{i+1}(s_t)] &= p_{i+1}(attack\vert s_t) . [r(s_t, a_t, s'_{t+1}) + \gamma V^{\pi_i}(s'_{t+1}] \\
  &\quad + [p_{i+1}(s_t) - p_{i+1}(attack\vert s_t)].[r(s_t, a_t, s_{t+1)}) + \gamma V^{\pi_i}(s_{t+1}]\\
  &\quad - V^{\pi_i}(s_t)
\end{aligned}
\end{equation}
where $p_{i+1}(s_t)$ is the probability of choosing an experience beginning with either nominal or crafted state $s_t$ form the experience memory in the $i+1$th iteration, and $p_{i+1}(attack\vert s_t) = p_{i+1}(s_t) - p^{nominal}_{i+1} (s_t)$ is the probability of choosing an experience sample beginning with an adversarially-crafted state $s_t$. It is noteworthy that adversarial perturbations add bias to the expected TD-error. It can be seen that, for the effect of this bias to be decreasing as $i$ increases, the following condition must hold true:
\begin{equation}
    p_{i+1}(s_t) - p_{i+1} (attack\vert s_t) > p_{i}(s_t) - p_i(attack\vert s_t) 
\end{equation}
That is, the probability of sampling nominal experiences starting with $s_t$ from the experience memory must be increasing with $i$. In the case of a state-neutral adversary, and assuming the uniform sampling from experiences, this condition reduces to:
\begin{equation}
    \forall s_t\in S: p^{nominal}_{i+1}(s_t) > p(attack)
\end{equation}
Which can be interpreted as $p(attack) < 0.5$. This is in agreement with the results reported in \cite{behzadan2017whatever} for non-contiguous, non-targeted adversarial example attacks against DQN agents.



\subsection{Experimental Results}
\label{sec:ch5res}
To evaluate the practical implications of the theoretical analyses of this section, we study the training performance of a CartPole DQN policy under non-targeted attacks with perturbation probabilities of $0.2, 0.4, 0.8, \text{and }1.0$. In these experiments, the attacks begin after the convergence of the policy to optimal performance. 

\begin{figure}[H]
	
	\centering
	
	\includegraphics[width = 0.5\columnwidth]{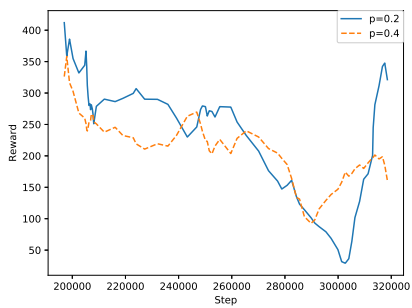}
	
	\caption{Training Performance Under Non-Targeted Attack with p(attack)= 0.2 and p(attack) = 0.4}
	
	\label{fig:Att02}
	
\end{figure}

\begin{figure}[H]
	
	\centering
	
	\includegraphics[width = 0.5\columnwidth]{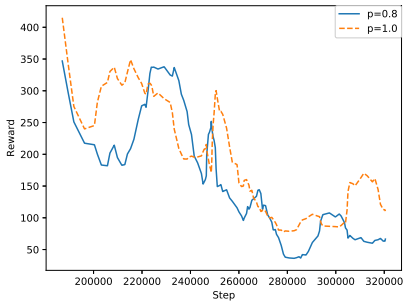}
	
	\caption{Training Performance Under Non-Targeted Attack with p(attack)= 0.8 and p(attack) = 1.0}
	
	\label{fig:Att08}
	
\end{figure}

The results are presented in figures \ref{fig:Att02} and \ref{fig:Att08}. It can be seen that for $p(attack) = 0.2$ and $p(attack) = 0.4$, the training process recovers rather quickly. However, for $p(attack) = 0.8$ and $p(attack) = 1.0$, the recovery fails to realize within the observed training horizon. It is noteworthy that the early peaking observed in Figure \ref{fig:Att08} are due to residual unperturbed experiences still remaining in the replay memory, the impact of which immediately fades at around 50000 steps after the attack begins, which is equivalent to the number of experiences required to completely overwrite the memory. 

\section{Adversarially-Guided Exploration Mechanism for Sample-Efficient Adversarial Training}
There exists a noteworthy difference between the theoretical adversaries considered so far and one that crafts perturbations through adversarial examples. As reported in \cite{pinto2017robust} and \cite{behzadan2017whatever}, training on adversarial examples enhances the resilience of the policy to perturbations crafted using the same technique. Similar to the case of adversarial training for deep learning classifiers \cite{shaham2018understanding}, this phenomenon can be explained from the perspective of regularization: adversarial example perturbations of states provide the means for regularization of the policy (or value function) through data augmentation. Therefore, training the policy over adversarial examples of states generated with a certain attack mechanism results in the enhancement of resilience and robustness of the policy to perturbations crafted via that mechanism. 

However, current procedures for training over adversarial examples (e.g., \cite{pinto2017robust}\cite{pattanaik2018robust} are based on ``blanket perturbation'', in which all state have an equal probability of being perturbed during training, thus leading to the deterioration of sample efficiency in DRL training. To alleviate this adverse effect, we propose the Adversarially-Guided Exploration (AGE) mechanism, which efficiently reduces the number of perturbed observations required to produce similar or better improvements in robustness compared to the results achieved by previous techniques. The proposed mechanism is based on the fact that not all states are equal with respect to the total regret produced by their perturbation. To account for this fact, the proposed AGE mechanism extends the classical $\epsilon$-greedy exploration mechanism by adjusting the probability of sampling actions for each state according to the \emph{adversarial state-action significance}, defined as follows: In the $(i+1)$th training iteration, the adversarial significance of any action $a$ in state $s$, denoted by $\zeta_{adv}^{\pi_i}(s, a)$, measures the maximum achievable adversarial gain, determined by the difference between maximum $Q$-value at state $s$ and $Q^{\pi_i}(s,a)$ with respect to actions. We define $\zeta_{adv}$ as the ratio of this difference to the sum of this difference for all actions $a\in A$. Furthermore, to retain the GLIE (Greedy in the Limit with Infinite Exploration) criteria of the $\epsilon$-greedy mechanism \cite{sutton1998reinforcement}, we formulate $\zeta_{adv}$ in the form of the Boltzmann probability\cite{cesa2017boltzmann}, with $\epsilon$ as the decaying temperature factor. Consequently, the formal definition of $\zeta_{adv}$ is as follows: 
\begin{equation}
    \zeta_{adv}^{\pi_i}(s, a) = \frac{\exp{(\max_{a'} Q^{\pi_i}(s, a') - Q^{\pi_i}(s, a)}/\epsilon)}{\sum_{\alpha \in A} \exp{(\max_{a'} Q^{\pi_i}(s, a') - Q^{\pi_i}(s, \alpha)}/\epsilon)}
\end{equation}


Algorithm \ref{alg:exploration} presents the details of our proposed exploration mechanism:
\begin{algorithm}[H] 
\caption{Adversarially-Guided Exploration (AGE) for Adversarial Training} 
\label{alg:exploration} 
\begin{algorithmic} 
    \REQUIRE $Q^{\pi_i}$, action space $A$
    \STATE \textbf{function }Adversarial\_Exploration(Current state $s$, exploration probability $\epsilon$)
    \bindent
    \FOR{all $a\in A$}
        \STATE $\zeta_{adv}^{\pi_i}(s, a) = \frac{\exp{(\max_{a'} Q^{\pi_i}(s, a') - Q^{\pi_i}(s, a)}/\epsilon)}{\sum_{\alpha \in A} \exp{(\max_{a'} Q^{\pi_i}(s, a') - Q^{\pi_i}(s, \alpha)}/\epsilon)}$
    \ENDFOR
    \IF{$rand()\leq \epsilon$}
         \STATE Sample action according to $\zeta_{adv}^{\pi_i}$ to perform
    \ELSE
        \STATE Perform $\argmax_a Q^{\pi_i}(s,a)$
    \ENDIF
    \eindent
\end{algorithmic}
\end{algorithm}

\section{Experiment Setup}
\textbf{Environment and Target Policies:} To evaluate the performance of AGE in adversarial training, we study the training efficiency and adversarial resilience of a DQN policy in the CartPole environment in OpenAI Gym \cite{brockman2016openai}. Table \ref{CartPole} presents the specifications of the CartPole environment, and Table \ref{CartPoleDQN} provides the parameter settings of each target policy.

\begin{table}[H]
\centering
\label{CartPole}
\caption{Specifications of the CartPole Environment}
\begin{tabular}{|l|l|}
\hline
Observation Space & \begin{tabular}[c]{@{}l@{}}Cart Position {[}-4.8, +4.8{]}\\ Cart Velocity {[}-inf, +inf{]}\\ Pole Angle {[}-24 deg, +24 deg{]}\\ Pole Velocity at Tip {[}-inf, +inf{]}\end{tabular} \\ \hline
Action Space      & \begin{tabular}[c]{@{}l@{}}0 : Push cart to the left\\ 1 : Push cart to the right\end{tabular}                                                                                      \\ \hline
Reward            & +1 for every step taken                                                                                                                                                             \\ \hline
Termination       & \begin{tabular}[c]{@{}l@{}}Pole Angle is more than 12 degrees\\ Cart Position is more than 2.4\\ Episode length is greater than 500\end{tabular}                                    \\ \hline
\end{tabular}
\end{table}

\begin{table}[H]
\centering
\label{CartPoleDQN}
\caption{Parameters of DQN Policy}
\begin{tabular}{|l|l|}
\hline
No. Timesteps               & $10^5$                \\ \hline
$\gamma$                    & $0.99$                \\ \hline
Learning Rate               & $10^{-3}$             \\ \hline
Replay Buffer Size          & 50000                 \\ \hline
First Learning Step         & 1000                  \\ \hline
Target Network Update Freq. & 500                   \\ \hline
Prioritized Replay          & True                  \\ \hline
Exploration                 & Parameter-Space Noise \\ \hline
Exploration Fraction        & 0.1                   \\ \hline
Final Exploration Prob.     & 0.02                  \\ \hline
Max. Total Reward           & 500                   \\ \hline
\end{tabular}
\end{table}

\textbf{Adversarial Agent:} In these experiments, the adversarial agent is a DQN agent with the hyperparameters provided in Table \ref{AdvDQN}. We consider a homogeneous perturbation cost function for all state perturbations, that is $\forall s, a': c_{adv}(s, a') = c_{adv}$. For both the resilience and robustness measurements, we set $c_{adv} = 1$ (i.e., each perturbation incurs a cost of $1$ to the adversary). The training process is terminated when the adversarial regret is maximized and the 100-episode average of the number of adversarial perturbations is quasi-stable for 200 episodes. 

\begin{table}[H]
\centering
\label{AdvDQN}
\caption{Parameters of DQN Policy}
\begin{tabular}{|l|l|}
\hline
Max. Timesteps              & $10^5$                \\ \hline
$\gamma$                    & $0.99$                \\ \hline
Learning Rate               & $10^{-3}$             \\ \hline
Replay Buffer Size          & 50000                 \\ \hline
First Learning Step         & 1000                  \\ \hline
Target Network Update Freq. & 500                   \\ \hline
Experience Selection        & Prioritized Replay    \\ \hline
Exploration                 & Parameter-Space Noise \\ \hline
Exploration Fraction        & 0.1                   \\ \hline
Final Exploration Prob.     & 0.02                  \\ \hline
\end{tabular}
\end{table}
\subsection{Results}
\begin{figure}[H]
	
	\centering
	
	\includegraphics[width = 0.5\columnwidth]{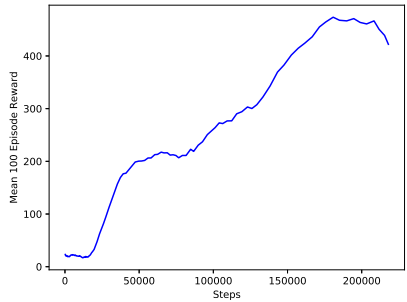}
	
	\caption{Training Performance of a CartPole DQN policy with AGE exploration}
	
	\label{fig:Explore1}
	
\end{figure}

\begin{figure}[H]
	
	\centering
	
	\includegraphics[width = 0.5\columnwidth]{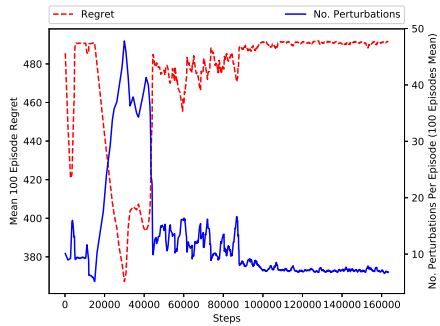}
	
	\caption{Adversarial Training Progress for Resilience Benchmarking of the DQN Policy with $\epsilon-$greedy exploration}
	
	\label{fig:ResDQN}
	
\end{figure}

\begin{figure}[H]
	
	\centering
	
	\includegraphics[width = 0.5\columnwidth]{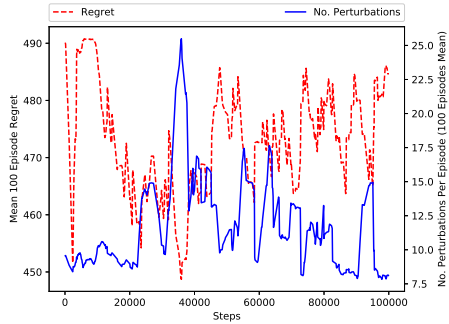}
	
	\caption{Training Performance of an Adversarial Agent Targeting the AGE-Trained Policy}
	
	\label{fig:Explore2}
	
\end{figure}

Figure \ref{fig:Explore1} illustrates the training performance of the DQN policy utilizing AGE for exploration. It can be seen than the training has successfully converged, and the progress is noticeably more stable than that of a DQN policy with $\epsilon$-greedy exploration. Furthermore, Figure \ref{fig:Explore2} depicts the training performance of a DQN-based adversarial resilience agent with the same configuration as presented in [Awaiting Appearance in Arxiv\footnote{http://www.vbehzadan.com/drafts/RobustBenchmark.pdf}]. In comparison with the performance of the same agent against the same policy trained using $\epsilon$-greedy exploration (Figure \ref{fig:ResDQN} ), two significant differences are observed: first, the adversarial agent targeting the AGE-trained policy achieves a lower regret and higher perturbation count in the same number of training iterations as its counter-part. Second, the training process targeting the AGE-trained policy fails to converge in 100000 iterations, whereas its counter-part converged at around 90000 iterations. These results indicate the superior resiliency of the AGE-trained policy over the nominal policy, thereby verifying the effectiveness of AGE in improving the adversarial resilience of policies.

Furthermore, in comparison with to the best-case scenario of adversarial training of the nominal DQN policy (as presented in Figure \ref{fig:Att02}), it can be seen that the AGE-based training process requires significantly fewer samples for convergence. This comparison further verifies the efficiency of our proposed scheme with respect to sample complexity. 

\section{Conclusion}
This paper formally establishes the limits of adversarial training in DQN agents with respect to the ratio of perturbed training experience to the nominal (i.e., unperturbed) experiences. We then address the sample-inefficiency of current adversarial training techniques, and present the Adversarially-Guided Exploration (AGE) mechanism to improve upon this shortcoming. The presented experimental results demonstrate the feasibility of this exploration mechanism in comparison with the traditional decaying $\epsilon$-greedy and parameter-space noise exploration algorithms.%
%
%
 \bibliographystyle{splncs04}
 \bibliography{ref}
%




\end{document}